\documentclass[letterpaper]{article} 
\usepackage{aaai24}  
\usepackage{times}  
\usepackage{helvet}  
\usepackage{courier}  
\usepackage[hyphens]{url}  
\usepackage{graphicx} 
\usepackage{natbib}  
\usepackage{caption} 
\usepackage{graphicx}
\usepackage{amsmath}
\usepackage{amssymb}
\usepackage{booktabs}
\usepackage{enumitem}
\usepackage{soul}
\usepackage{color}
\usepackage{bm}

\usepackage{algorithm}
\usepackage{algorithmic}

\usepackage{newfloat}
\usepackage{listings}
\title{Cap2Aug: Caption guided Image to Image data Augmentation}

\author{
    Aniket Roy\textsuperscript{\rm 1},
    Anhsul Shah$^*$\textsuperscript{\rm 1},
    Ketul Shah$^*$\textsuperscript{\rm 1},
    Anirban Roy\textsuperscript{\rm 2},
    Rama Chellappa\textsuperscript{\rm 1}
}

\affiliations{
    \textsuperscript{\rm 1} Johns Hopkins University
     \textsuperscript{\rm 2} SRI International
}

\usepackage{bibentry}

\begin{document}

\maketitle

\begin{abstract}
Visual recognition in a low-data regime is challenging and often prone to overfitting. To mitigate this issue, several data augmentation strategies have been proposed. However, standard transformations, e.g., rotation, cropping, and flipping provide limited semantic variations. To this end, we propose Cap2Aug, an image-to-image diffusion model-based data augmentation strategy using image captions as text prompts. We generate captions from the limited training images and using these captions edit the training images using an image-to-image stable diffusion model to generate semantically meaningful augmentations. This strategy generates augmented versions of images similar to the training images yet provides semantic diversity across the samples. We show that the variations within the class can be captured by the captions and then translated to generate diverse samples using the image-to-image diffusion model guided by the captions. However, naive learning on synthetic images is not adequate due to the domain gap between real and synthetic images. Thus, we employ a maximum mean discrepancy (MMD) loss to align the synthetic images to the real images for minimizing the domain gap. We evaluate our method on few-shot and long-tail classification tasks and obtain performance improvements over state-of-the-art, especially in the low-data regimes.
\end{abstract}

\def\thefootnote{*}\footnotetext{Equal contribution}\def\thefootnote{\arabic{footnote}}


\section{Introduction}
\label{sec:intro}

\begin{figure}[t]
  \centering
   \includegraphics[scale=0.55]{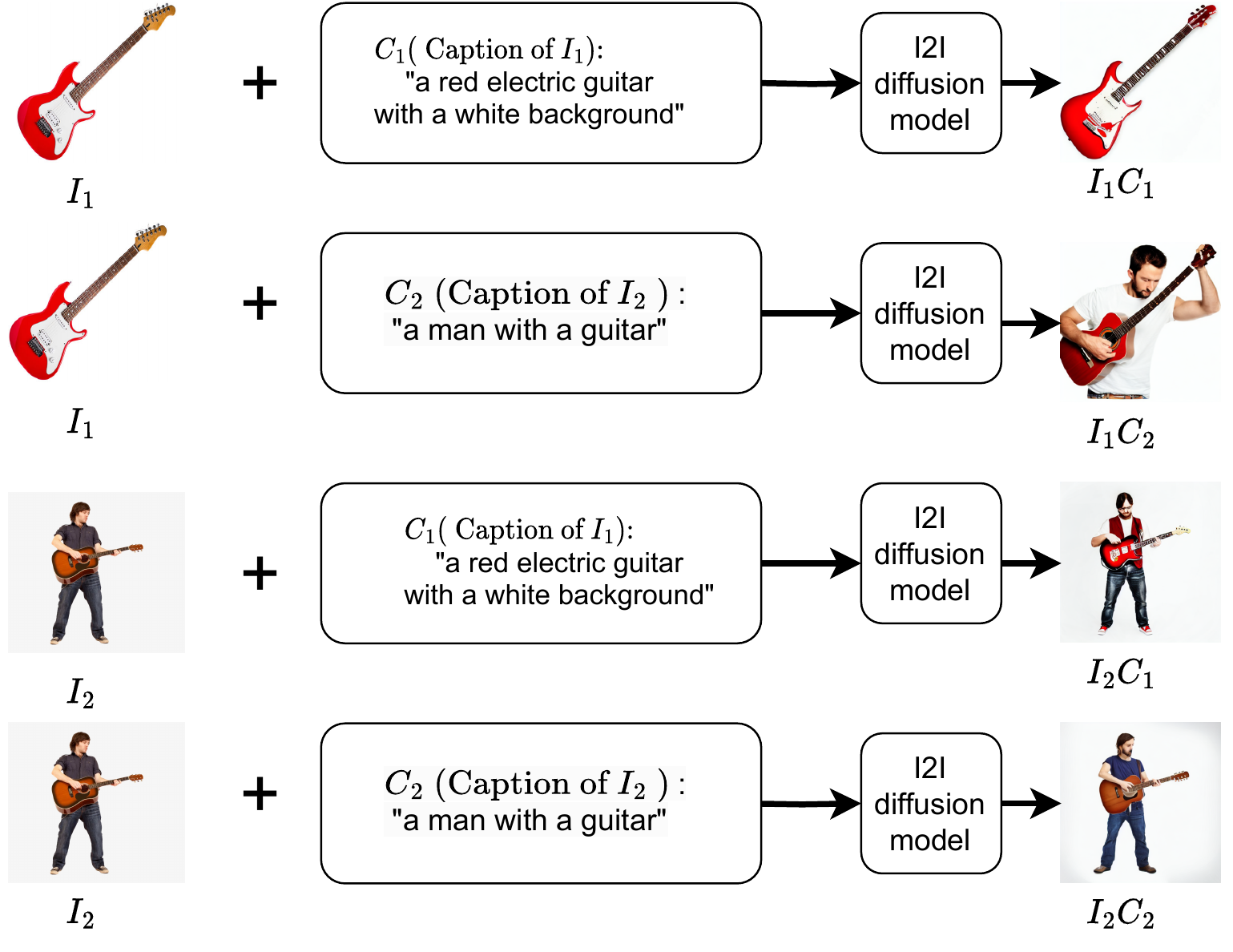}
   \caption{Idea of Cap2Aug: real images $I_1$ (guitar) and $I_2$ (person playing guitar) are fed to a captioning model to generate ``a red electric guitar with a white background'' ($C_1$) and ``a man with a guitar'' ($C_2$) as captions, respectively. Image $I_1$ and caption $C_1$ when fed to an image-to-image diffusion model generates a synthetic image $I_1C_1$ - the image of a guitar similar to $I_1$ with minor changes (fine changes in guitar head, body). On the other hand, when image $I_1$ and caption $C_2$ are fed to the image-to-image diffusion model, it generates a man with a guitar in his hand ($I_1C_2$). This is replicated using image $I_2$ for generating synthetic images $I_2C_1$ and $I_2C_2$ respectively using captions $C_1$ and $C_2$.}
   \vspace{-0.3cm}
   \label{fig:teaser_fig}
\end{figure}

Supervised image classification approaches have achieved near-human performance \cite{AlexNet,He_2016_CVPR} by leveraging large-scale datasets~\cite{ImageNet,cao2021concept}. However, learning from limited data remains challenging, especially in few-shot setups, where only 1-5 samples are available for each class. To address this challenge, existing approaches consider various data augmentation approaches to expand the training set. For example, Jian et al.~\cite{Jian2022LabelHalluc} generate pseudo labels for the base class samples and use these samples for increasing the number of novel class samples. Assoalign~\cite{Afrasiyabi_2020_ECCV} uses base-class samples in addition to the novel class samples to generate new samples in an adversarial framework. Roy et al.~\cite{roy2022felmi} use hard-mixup to combine existing samples to generate additional samples. However, these augmentation techniques such as rotation, scaling, flipping, etc. generate images with limited semantic variations.


Recently, generative models, such as DALL-E~\cite{ramesh2022hierarchical} and stable diffusion~\cite{rombach2022high}, are shown to be successful in generating realistic images. Moreover, vision and language models, such as CLIP~\cite{radford2021learning} and BLIP~\cite{li2022blip}, are shown to generate diverse human-like captions from images. Inspired by the recent success of these generative models, we propose Cap2Aug - a data augmentation strategy that provides semantic variations aided by generative models. We first generate a set of captions from an image using a captioning model and then generate novel images based on the captions using an image-to-image diffusion model with captions as text prompts. While generating these images, captions from one image can be used as text prompts for another image to ensure semantic diversity in novel images. Finally, the classifier can be trained with the augmented dataset. The Cap2Aug framework is shown in Fig.~\ref{fig:teaser_fig}. However, directly combining synthetic images with real images may not be effective due to the domain gap between these images. Thus, we propose a maximum mean discrepancy (MMD) loss~\cite{long2015learning} to align the features of the synthetic images to real images.

As Cap2Aug considers generative models to augment images and these models are usually trained on large-scale datasets. Thus, our approach is not directly comparable to few-shot approaches~\cite{Jian2022LabelHalluc, roy2022felmi, afrasiyabi2021mixture, afrasiyabi2022matching} that do not consider external sources of images. Our goal is to develop a data-augmentation framework that leverages existing generative models. Thus, Cap2Aug can be compared to existing approaches that use additional datasets or models to improve the classification performance. We consider image classification in a few-shot and  long-tail data distribution to evaluate our approach.
In this context, our contributions include:

\begin{itemize}[topsep=0pt,noitemsep]

\item We propose Cap2Aug - a data augmentation strategy leveraging image-to-image generative models with image captions as text prompts.
We have validated this approach for long-tail and few-shot classification tasks.
Cap2Aug is particularly suitable for few-shot image classification where only a few training images are available.
\item We propose an MMD-based loss function to align synthetic images to real images in order to reduce the domain gap between real and synthetic images.
\item We validate our approach on standard long-tail classification on ImageNet-LT and eleven few-shot classification benchmarks and obtain significant improvements over the state-of-the-art. 

\end{itemize}

\section{Related Work}
\label{sec:related_work}

Traditionally, meta-learning-based methods have been useful for few-shot learning problems, e.g., 
Prototypical Networks~\cite{ProtoNet}, Relation Networks~\cite{Sung_2018_CVPR}, Matching networks~\cite{MatchNet}, TADAM~\cite{TADAM}, MAML~\cite{MAML}, LEO~\cite{rusu2018metalearning}, 
MetaOptNet~\cite{Ravichandran_2019_ICCV} etc.
However, instead of traditional meta-learning approaches, recent methods rely on simple yet efficient transfer-learning approach.
RFS~\cite{tian2020rethink} shows that contrastive pre-training on the large base dataset and simple finetuning on the novel examples outperforms all the meta-learning baselines.
SKD~\cite{rajasegaran2020selfsupervised} added rotational self-supervised distillation to further improve the performance.
Using complementary strengths of invariant and equivariant representations and self-distillation, Rizve et al.~\cite{rizve2021exploring} performs significantly better than previous few-shot learning methods. 
Feature-level knowledge distillation using partner-assisted learning~\cite{ma2021partner} has proven to be effective.
Recent few-shot classification techniques use - co-adaptation of discriminative features~\cite{chikontwe2022cad}, mutual centralized learning~\cite{liu2022learning}, discriminative subspace~\cite{zhu2022ease}, multi-task representation learning~\cite{bouniot2022improving}, contrastive learning~\cite{yang2022few}, task-aware dynamic kernels~\cite{ma2022learning}, CLIP adaptation~\cite{zhang2022tip} and large pretrained networks~\cite{hu2022pushing}.



\textbf{Data augmentation for few-shot learning.}
A few recent methods use  readily available base datasets in addition to novel class samples.
AssoAlign~\cite{Afrasiyabi_2020_ECCV} selects the nearest neighbors of the novel class samples from the abundant base dataset and adversarially aligns those for training. Jian et al.~\cite{Jian2022LabelHalluc} generate pseudolabels for the  base dataset using a classifier trained on the novel classes. Roy et al.~\cite{roy2022felmi} use hard-mixup sample selection to further improve the performance. Afrasiyabi et al.~\cite{afrasiyabi2021mixture} consider mixture-based feature space learning~\cite{afrasiyabi2021mixture} and matching feature sets~\cite{afrasiyabi2022matching}.
However, these approaches have not used the text label information. 

\textbf{Multi-modal few-shot learning.} 
Semantic information is useful for few-shot classification~\cite{afham2021rich}.  
Padhe et al.~\cite{pahde2021multimodal} use multi-modal prototypical networks for few-shot classification and Yang et al.~\cite{yang2022sega} utilize semantic guided attention to integrate the rich semantics into few-shot classification. Xu et al.~\cite{xu2022generating} generates representative samples for few-shot learning using text-guided variational autoencoder. 
Wang et al.~\cite{wang2022multi} uses multi-directional knowledge transfer for multi-modal few-shot learning. 
Text-guided prototype completion~\cite{zhang2021prototype} also helps few-shot classification. 

\textbf{Vision-language models.}
Recent advancements in large-scale vision language pretrained models enable significant improvements in multi-modal learning with CLIP~\cite{radford2021learning}, GPT-3~\cite{brown2020language}, DALLE~\cite{ramesh2022hierarchical}, stable diffusion~\cite{rombach2022high} etc. Diffusion models are state-of-the-art text-to-image generative models~\cite{ho2020denoising, nichol2021glide, ramesh2022hierarchical, rombach2022high}, which are trained on large-scale image and text corpus and produces surprisingly well images just from texts.
Vision-language pretraining model CLIP~\cite{radford2021learning} helps to improve zero-shot performance across several datasets. Prompt tuning method CoOp~\cite{zhou2022learning} optimizes learnable prompts for better few-shot adaptation. CoCoOp~\cite{zhou2022conditional} and VT-CLIP~\cite{zhang2021vt} used a text-conditioned intermediate network for joint image-text training. CLIP-adapter~\cite{gao2021clip} tried to adapt the powerful CLIP features with a lightweight residual style network adapter for few-shot adaptation. Tip-Adapter~\cite{zhang2022tip} extend this using a training free key-value based cache model and obtained a performance boost. CALIP~\cite{guo2022calip} propose to use parameter-free attention to elevate CLIP performance in both zero-shot and few-shot settings. SuS-X~\cite{udandarao2022sus} extends Tip-adapter using image-text distance and dynamic support set. 

\textbf{Diffusion models.} Diffusion models~\cite{ho2020denoising, nichol2021glide, ramesh2022hierarchical, rombach2022high} are text-to-image generative models that generate realistic images from free-form texts. Compared to other generative models, e.g, GANs~\cite{goodfellow2020generative}, flow-based models~\cite{kobyzev2020normalizing}, diffusion models produce more diverse and realistic images.  In this work, we consider the stable diffusion~\cite{rombach2022high} a text-guided image-to-image diffusion model. The goal is to update the input image according to the given text prompt. In a diffusion model, Gaussian noise is gradually added to an input image and a denoising autoencoder is trained to reverse the noising process. To learn the data distribution, the denoising autoencoder minimizes the variational lower bound of the data distribution. Due to the Gaussian noise assumption, the loss can be expressed in a closed form and efficient sampling strategies can be utilized. 
Instead of applying the diffusion process on image space, Rombach et al.~\cite{rombach2022high} propose to project the high-dimensional input to low-dimensional latent space for fast and efficient computation. For text-guided image generation, a text-specific encoder is used that projects texts to an intermediate representation and is jointly trained with the denoising autoencoder.

\section{Proposed Approach}
\label{sec:method}

\begin{figure}[t]
  \centering
   \includegraphics[scale=0.45]{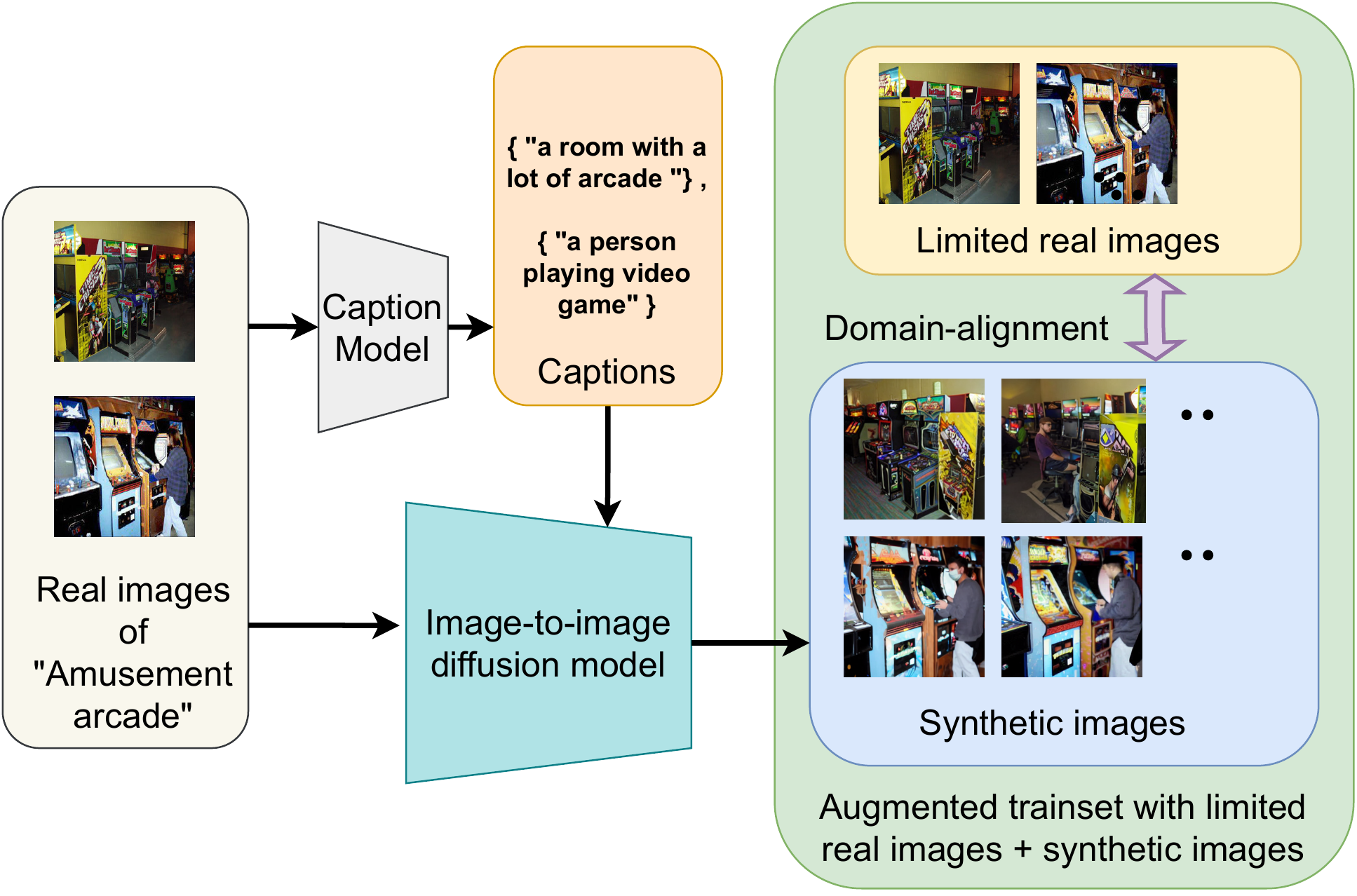}
   \caption{Overview of Cap2Aug. We generate captions from the real images using the BLIP caption model~\cite{li2022blip}. The generated captions and real images are fed to the image-to-image diffusion model~\cite{rombach2022high} to generate plenty synthetic images. We then align the synthetic images with the real images to reduce the domain gap. Finally, the combined set of limited real images and aligned synthetic images are used to learn a classifier for the novel class.}
   \label{fig:teaser}
\end{figure}

In this section, we describe our proposed text-guided data augmentation strategy. The proposed method has four main components: 
1) Generate captions from the images using a caption model,
2) Generate synthetic augmentations using text guided image-to-image diffusion model, where captions are provided as text prompts,
3) Cache-based adapter on top of the CLIP model,
4) Domain alignment between real and synthetic images using MMD.

\begin{figure}[t]
  \centering
   \includegraphics[scale=0.5]{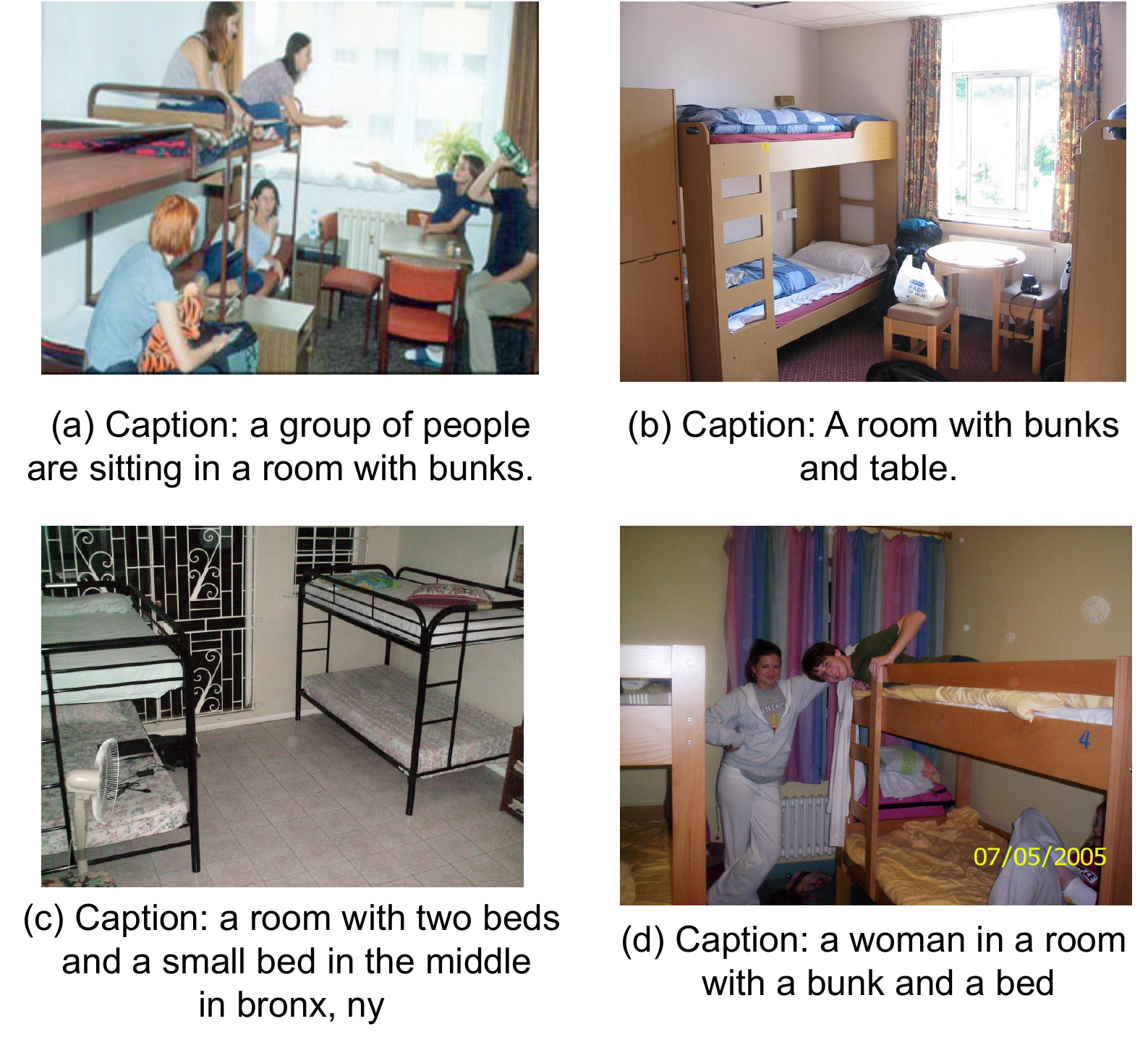}
   \caption{Illustration of image captioning using BLIP model: Images from the class ``youth hostel''. Several images capture the characteristics of the class ``youth hostel''. E.g., images contain a bunk bed and a group of people sitting as shown in the captions generated from the images. }
   \label{fig:blip_caption}
\end{figure}

\subsection{Caption generation}

Captions capture the semantic information of  images with succinct texts. Current large-scale vision and language  methods, e.g., CLIP~\cite{radford2021learning}, BLIP~\cite{li2022blip} achieve impressive performance in image captioning. We use the bootstrapping language-image pretraining (BLIP)~\cite{li2022blip} in this work. 

We capture the diversity and domain information residing in the few training examples by captioning the images using the off-the-shelf BLIP caption model~\cite{li2022blip}. For an N-way K-shot few-shot setup, we have $I_{NK}$ training images containing $K$ images each for $N$ classes. Now, using the BLIP caption model, we generate captions $C_{i}$ for each image $I_{i}$ in the training set. 
\begin{equation}
    C_{i} = \textit{BLIP-model}(I_{i})
\end{equation}
Captions for each class contain class-dependent semantic information with diversity across samples. E.g., In Sun397 dataset~\cite{xiao2010sun}, there exists a class ``youth hostel'', which contains images of a group of people sitting on a bed and a couple of bunk beds as shown in Fig.~\ref{fig:blip_caption}. These are typical characteristics of the ``youth hostel'' class. Therefore, generated captions capture the semantic characteristics of the class information as shown in Fig.~\ref{fig:blip_caption}. 

\subsection{Caption guided image to image translation}
Traditional image augmentation methods rely on fixed transformations e.g., translation, rotation, etc. To the contrary, we attempt to generate an augmented version of images using an image-to-image diffusion model by editing these images using captions.

Stable diffusion is a diffusion model conditioned on text embedding of CLIP ViT-L/14~\cite{radford2021learning} text encoder and trained on LAION-400M dataset~\cite{rombach2022high} of image-text pairs. The stable diffusion model can generate  realistic images from text descriptions.
In this work, the stable diffusion model is conditioned on the text prompts that are based on the diffusion-denoising mechanism proposed by SDEdit~\cite{meng2021sdedit}. The method generates/updates images by iterative denoising through a stochastic differential equation conditioned on the encoded version of the text prompt. 
Examples of the input image and caption pairs and corresponding generated images are shown in Fig.~\ref{fig:teaser_fig}.

Now, the augmented versions of the images are generated by,
\begin{equation}
    I_{i}C_{j} = \textit{I2I}(I_{i}, C_{j}) \quad for \qquad i,j = 1, .., K
\end{equation}

where $I2I$ is the pretrained image-to-image diffusion model, $I_{i}$ is the $i$-th image and the corresponding caption is denoted by $C_i$. In an N-way K-shot classification problem, for each class, we have K training images such as $I_1$, $I_2$,... $I_K$. The corresponding captions generated by the BLIP-caption model are $C_1$, $C_2$, ... $C_K$, respectively. Now, we can generate diverse images pairing $(I_i, C_j)$ denoted by $I_iC_j$ for $i,j=1, .., K$ as shown in Fig.~\ref{fig:teaser_fig}. 

Note that, generated images with self-caption viz., $(I_i, C_i)$ is denoted by $I_iC_i$, i.e., image-to-image generation using captions from the image itself would still result in some style difference in the image. In Fig.~\ref{fig:teaser_fig} ($I_1C_1$), image-to-image translation of the ``guitar image'' with its own caption (i.e., ``a red electric guitar with a white background''), still generates an image with a different semantic content (i.e., difference in the guitar head). Hence, this can also be considered as a useful augmentation.

More interesting and diverse images are generated by cross image-caption pairs $(I_i, C_j)$ ($i \neq j$), where the style of image caption $C_j$ is translated to generated images from $I_i$ through image-to-image stable diffusion model. For instance, an image of a guitar is translated to a person playing a guitar using the caption as ``a man with a guitar'' as shown in Fig.~\ref{fig:teaser_fig} ($I_2C_1$).

We generate augmented versions of the training images conditioned on the class information captured by captions. 
Our objective is to provide semantic variations of the existing training images, not generating new samples itself using the off-the-shelf generative models.
Note that, we are not explicitly using the class labels for generating the images. Since the diffusion models are trained on large-scale datasets, therefore generating images using class labels might violate the inherent problem of low-data regime e.g., long-tail or few-shot setting. 

\subsection{Cache based adapter}
In a low-data regime, e.g., few-shot classification, in addition to the classifier we used a cache-based adapter.
Predictions from cache-based adapters can be considered as a combination of pre-trained CLIP model prediction and few-shot finetuned model prediction.
After generating $K'$ number of synthetic images per class, we have $N(K+K')$ number of images in the training set.
Now, similar to Tip-Adapter~\cite{zhang2022tip}, we construct a key-value cache model for adapting CLIP visual encoder. 
We extract the CLIP features for both the real and synthetic images $F_{CLIP}$ and the corresponding one-hot labels as $L_{one-hot}$. The cache model ($f_{\theta}$) is a 2-layer MLP initialized with CLIP features as keys, and the one-hot labels as values.
Keeping the CLIP backbone frozen, the adapter keys are learned using cross-entropy loss to adapt from the few examples using cross-entropy loss.
\begin{equation}
    \mathcal{L}_{CE} = CE( f_{\theta}(F_{CLIP}(I_{N(K+K')})), L_{one-hot} )
\end{equation}

Suppose, CLIP features for the training images are $F_{train}$ and the corresponding one-hot labels are $L_{train}$ and CLIP features for the test images are $f_{test}$. In the inference stage of Tip-Adapter, the predicted logits are computed as follows~\cite{zhang2022tip}:
\begin{equation}
    \text{logits} = a \phi(f_{test}F^{T}_{train})L_{train} + f_{test}W^{T}_{c} 
\end{equation}

where $a$ denotes the residual ratio, $W_c$ is CLIP's text classifier, $\phi(x) = exp(-\beta(1-x))$, and $\beta$ is a hyper-parameter.

\begin{figure}
    \centering
    {\includegraphics[scale=0.25]{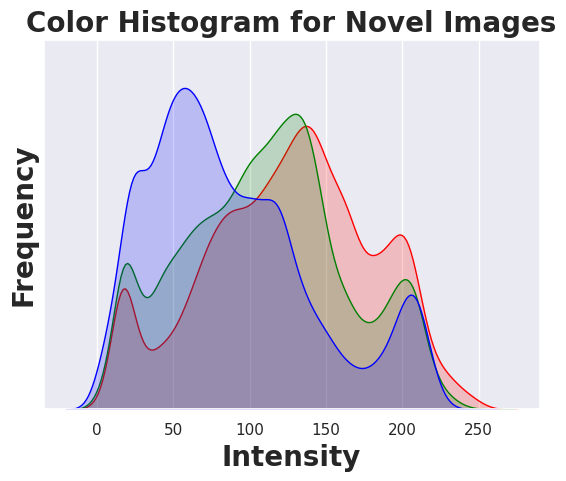}} 
    {\includegraphics[scale=0.25]{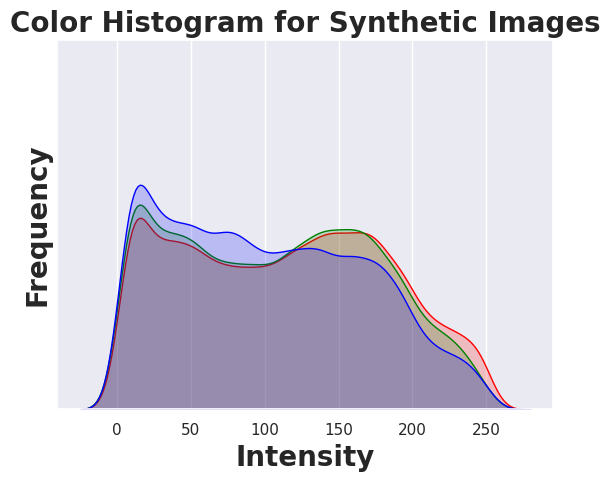}}
    \caption{Color histogram of novel and synthetic samples}
    \label{fig:color_histogram_novel_syn}
\end{figure}

\subsection{Domain alignment using MMD}
Despite the high-quality synthetic images, there exists a domain gap between the real and synthetic images in terms of the background, color and intensity distribution as shown in Fig.~\ref{fig:failure_cases}. The average color histograms of the real and synthetic images are shown in Fig.~\ref{fig:color_histogram_novel_syn}, which exhibits a distinction between these sets of images. To mitigate this, we use a multi-kernel Maximum Mean Discrepancy (MMD) loss to reduce the domain gap. MMD loss minimizes the domain gap by reducing the distance of the mean feature embeddings of the real and synthetic images.

Let's assume, given a source ($\mathcal{D}_s$) and the target domain ($\mathcal{D}_t$), samples are drawn from these domains with the distribution $P$ and $Q$, respectively over a set $\mathcal{X}$. The features of the samples from these domains are denoted as $\{z_{i}^s\}$ and $\{z_{i}^t\}$, respectively. A multi-kernel MMD ($D_{k}(P,Q)$) between probability distributions $P$ and $Q$ is defined as~\cite{long2015learning}: $D_k(P,Q) = || \mathbb{E}_{p} [\psi(z^s)] - \mathbb{E}_{q} [\psi(z^t)] ||_{\mathcal{H}_{k}}^{2}$
 where $k$ is the kernel function in the functional space, i.e., 
 $ k = \sum_{p=1}^{P} \alpha_{p} k_p$,
where $k_p$ is a single kernel. The feature map $\psi : \mathcal{X} \rightarrow \mathcal{H}_{k}$ maps into a reproducing kernel Hilbert space. 
$k = \{ \mathcal{N}(0, 0.5), \mathcal{N}(0, 1), \mathcal{N}(0, 2) \}$. If the kernel is \mbox{$k(x,y) = <\psi(x),\psi(y)>_{\mathcal{H}_k}$}, then using kernel trick, MMD can be estimated as without directly learning $\psi(\cdot)$ as:
\begin{multline}
     \bar{D}_k(P,Q) = \frac{1}{n_s^2} \sum_{i=1}^{n_s} \sum_{j=1}^{n_s} k(z_i^s, z_j^s) 
                    + \frac{1}{n_t^2} \sum_{i=1}^{n_t} \sum_{j=1}^{n_t} k(z_i^t, z_j^t) \\
                    - \frac{2}{n_s n_t} \sum_{i=1}^{n_s} \sum_{j=1}^{n_t} k(z_i^s, z_j^t)   
\end{multline}

Therefore, the MMD-loss between the real examples ($I_{NK}$) and synthetic examples ($I_{NK'}$) will be,

\begin{equation}
    \mathcal{L}_{MMD} = \bar{D}_k( f_{\theta}(I_{NK}), f_{\theta} (I_{NK'}) )
    \label{equ:loss_MMD}
\end{equation}

Finally, the adapter is trained with both losses.

\begin{equation}
    \mathcal{L} = \mathcal{L}_{CE} + \alpha * \mathcal{L}_{MMD}
    \label{equ:loss}
\end{equation}

The scaling parameter $\alpha$ is set experimentally and ablation on this parameter is shown in the experiments section.

\subsection{Implementation details}
For a fair comparison, we use CLIP~\cite{radford2021learning} with ResNet-50 as the visual encoder. On top of the feature extractor, the adapter is initialized as a 2-layer MLP  with cache keys as learnable parameters. We train the adapter using an AdamW optimizer with an initial learning rate of 0.001 with a cosine scheduler.
For generating image captions, we use the open-source implementation of BLIP-caption generator~\cite{li2022blip} provided in \texttt{diffusers} library from HuggingFace. We also use the same library for generating images using an image-to-image stable diffusion model with the “stable-diffusion-v1-5” model. More details are provided in the supplementary material.  
\section{Experiments}
\label{sec:experiments}

We evaluate the proposed approach on two tasks: 1) few-shot classification and 2) long-tail classification.

\subsection{Few-shot classification}

Data augmentation is more effective in data scarce regime. Hence, we validate our augmentation strategy for the few-shot classification task.
We perform few-shot experiments on eleven benchmark datasets - ImageNet~\cite{ImageNet}, Stanford-Cars~\cite{krause20133d}, UCF101~\cite{soomro2012ucf101}, Caltech-101~\cite{wah2011caltech}, Flowers102~\cite{nilsback2008automated}, SUN397~\cite{xiao2010sun}, DTD~\cite{cimpoi2014describing}, EuroSAT~\cite{helber2019eurosat}, FGVCAircraft~\cite{maji2013fine}, OxfordPets~\cite{parkhi2012cats}, and Food-101~\cite{bossard2014food}. We follow the protocol of Tip-Adapter~\cite{zhang2022tip} to train models with 2, 4, 8, and 16 shots and test on the full test set. Following standard practice, we consider classification accuracy as the metric. 
For a fair comparison, we make sure the number of images used in training are same for the baseline and in our approach. 

We compare our method with the state-of-the-art Tip-Adapter~\cite{zhang2022tip} and CoOp~\cite{zhou2022learning}, in Table.~\ref{table:imagenet_fs}, Table.~\ref{tab:eurosat_sun_ucf}, Table.~\ref{tab:cars_food_dtd}, Table.~\ref{tab:pets_flowers_fgvc} and, Table.~\ref{table:caltech101}, for few-shot classification tasks on eleven different benchmarks. Our method consistently outperforms the state-of-the-art in most cases including the challenging fine-grained classification datasets. 

\begin{table}
  \small
  \centering
  \caption{Comparison on ImageNet}
  \vspace{-0.4cm}
  \scalebox{0.8}{
  \begin{tabular}{lccccc}
  \toprule
  {\bf Method} & {\bf 2-shot} & {\bf 4-shot} & {\bf 8-shot} & {\bf 16-shot}\\ 
  \midrule
 {Tip~\cite{zhang2022tip} } & {60.96} & {60.98} & {61.45} & {62.03} \\
 {CoOp~\cite{zhou2022learning}} & {50.88} & {56.22} & {59.93} & {62.95} \\
 {Tip-F~\cite{zhang2022tip} } & {61.69} & {62.52} & {64.00} & {65.51} \\
 {Ours} & {\textbf{62.72}} & {\textbf{63.50}} & {\textbf{64.95}} & {\textbf{66.32}} \\  	  
\bottomrule
  \end{tabular}}
  \label{table:imagenet_fs}
\end{table}

\begin{table}[!ht]
  \centering
  \caption{Ablation of Number of Synthetic images (K) on ImageNet}
    \vspace{-0.4cm}
  \scalebox{0.7}{
  \begin{tabular}{lccccc}
  \toprule
  {\bf K } & {\bf 4} & {\bf 16} & {\bf 40} & {\bf 80}\\ 
  \midrule
 {2-shot} & {62.7} & {63.1} & {63.6} & {64.2} \\
 {4-shot} & {63.0} & {63.5} & {63.9} & {64.5} \\
 {8-shot} & {63.4} & {64.1} & {64.9} & {65.6} \\
 {16-shot} & {64.1} & {64.6} & {65.7} & {66.3} \\	  
\bottomrule
  \end{tabular}}
    \vspace{-0.4cm}
  \label{table:ablation_no_syn_img}
\end{table}

\subsection{Long-tail classification}
Long-tail classification has both data-scarce and data-abundant classes, therefore is a good test case for validating our data augmentation strategy.
We conduct experiments on large-scale long-tailed ImageNet-LT benchmark and obtain performance improvements over SOTA~\cite{tian2022vl} using Cap2Aug data augmentation as shown in Tab.~\ref{table:imagenet_lt}. We provide results for overall accuracy, many-shot (100 samples), medium-shot (20-100 samples), and few-shot (20 samples) cases. 
In this experiment, 40 images are generated for all the classes and used those as augmented data. We observe the performance gain is higher for few-shot classes in Tab.~\ref{table:imagenet_lt}.

\begin{table}[!ht]
  \centering
  \caption{\small{Comparison on ImageNet-LT}}
  \vspace{-0.4cm}
  \scalebox{0.6}{
  \begin{tabular}{lccccc}
  \toprule
  {\bf Method} & {\bf Overall Acc.} & {\bf Many-shot} & {\bf Medium-shot} & {\bf Few-shot}\\ 
  \midrule
  {ResLT~\cite{cui2022reslt}} & {55.1} & {63.3} & {53.3} & {40.3} \\
  {PaCo~\cite{cui2021parametric}} & {60.0} & {68.2} & {58.7} & {41.0} \\
  {LWS~\cite{kang2019decoupling}} & {51.5} & {62.2} & {48.6} & {31.8} \\
  {Zero-shot CLIP~\cite{radford2021learning}} & {59.8} & {60.8} & {59.3} & {58.6} \\ 
 {DRO-LT~\cite{samuel2021distributional} } & {53.5} & {64.0} & {49.8} & {33.1} \\
 {VL-LTR~\cite{tian2022vl}} & {70.1} & {77.8} & {67.0} & {50.8} \\
 {Ours} & {\textbf{70.9}} & {\textbf{78.5}} & {\textbf{67.7}} & {\textbf{51.9}} \\  	  
\bottomrule
  \end{tabular}}
  \vspace{-0.4cm}
  \label{table:imagenet_lt}
\end{table}

\begin{table}[!ht]
  \centering
  \caption{Various backbones for ImageNet 16-shot classification}
    \vspace{-0.4cm}
  \scalebox{0.7}{
  \begin{tabular}{lccccc}
  \toprule
  {\bf Method} & {\bf RN50} & {\bf RN101} & {\bf ViT/32} & {\bf ViT/16}\\ 
  \midrule
  {Tip-F~\cite{zhang2022tip}} & {65.51} & {68.56} & {68.65} & {73.69} \\
  {Ours} & {\textbf{66.32}} & {\textbf{69.20}} & {\textbf{69.70}} & {\textbf{74.70}} \\  	  
  \bottomrule
  \end{tabular}}
  \vspace{-0.4cm}
  \label{table:backbone}
\end{table}

\begin{table*}
  \centering
  \caption{Comparison on EuroSAT, SUN397 and UCF101}
  \vspace{-0.4cm}
  \scalebox{0.7}{
  \begin{tabular}{@{}lccccccccccccc@{}}
    \toprule
    {} & \multicolumn{4}{c}{\textbf{EuroSAT}} & \multicolumn{4}{c}{\textbf{SUN397}} & \multicolumn{4}{c}{\textbf{UCF101}} \\
    \cmidrule(lr){2-5} \cmidrule(lr){6-9} \cmidrule(lr){10-13}
    {\textbf{Shots}} & {2} & {4} & {8} & {16} & {2} & {4} & {8} & {16} & {2} & {4} & {8} & {16}  \\
    \toprule 
    {Tip~\cite{zhang2022tip} } & {61.68} & {65.32} & {67.95} & {70.50} & {62.70} & {64.15} & {65.62} & {66.85} & {64.74} & {66.46} & {68.68} & {70.58} \\
    {CoOp~\cite{zhou2022learning}} & {61.50} & {70.18} & {76.73} & {83.53} & {59.48} & {63.47} & {65.52} & {69.26} & {64.09} & {67.03} & {71.92} & {75.71} \\ 
    {Tip-F~\cite{zhang2022tip}} & {66.15} & {74.12} & {77.3} & {82.54} & {63.64} & {66.21} & {68.87} & {70.47} & {66.43} & {70.55} & {74.01} & {77.03} \\
    {Ours} & {\textbf{67.03}} & {\textbf{77.37}} & {\textbf{77.5}} & {\textbf{83.64}} & {\textbf{64.605}} & {\textbf{67.47}} & {\textbf{68.93}} & {\textbf{70.9}} & {\textbf{68.57}} & {\textbf{71.76}} & {\textbf{74.12}} & {\textbf{77.24}} \\  	
\bottomrule
  \end{tabular}}
  \label{tab:eurosat_sun_ucf}
\end{table*}

\begin{table*}
  \centering
  \caption{Comparison on OxfordPets, OxfordFlowers and FGVC}
  \vspace{-0.4cm}
  \scalebox{0.7}{
  \begin{tabular}{@{}lccccccccccccc@{}}
    \toprule
    {} & \multicolumn{4}{c}{\textbf{OxfordPets}} & \multicolumn{4}{c}{\textbf{OxfordFlowers}} & \multicolumn{4}{c}{\textbf{FGVC}} \\
    \cmidrule(lr){2-5} \cmidrule(lr){6-9} \cmidrule(lr){10-13}
    {\textbf{Shots}} & {2} & {4} & {8} & {16} & {2} & {4} & {8} & {16} & {2} & {4} & {8} & {16}  \\
    \toprule 
    {Tip~\cite{zhang2022tip} } & {87.03} & {86.45} & {87.03} & {88.14} & {79.13} & {83.80} & {87.98} & {89.89}  & {21.21} & {22.41} & {25.59} & {29.76} \\
    {CoOp~\cite{zhou2022learning}} & {82.64} & {86.70} & {85.32} & {87.01} & {77.5} & {86.20} & {91.18} & {94.51} & {18.68} & {21.87} & {26.13} & {31.26}\\
    {Tip-F~\cite{zhang2022tip}} & {87.03} & {87.54} & {88.09} & {89.70} & {82.30} & {85.83} & {90.51} & {94.80} & {23.19} & {24.80} & {29.21} & {34.55} \\
    {Ours} & {\textbf{87.33}} & {\textbf{87.92}} & {\textbf{88.20}} & {\textbf{89.725}} & {\textbf{83.06}} & {\textbf{86.64}} & {\textbf{91.44}} & {94.51}  & {\textbf{23.76}} & {\textbf{24.87}} & {\textbf{29.82}} & {34.38} \\   	
\bottomrule
  \end{tabular}}
  \label{tab:pets_flowers_fgvc}
\end{table*}

\begin{table*}
  \centering
  \caption{Comparison on StanfordCars, Food101 and DTD}
  \vspace{-0.4cm}
  \scalebox{0.7}{
  \begin{tabular}{@{}lccccccccccccc@{}}
    \toprule
    {} & \multicolumn{4}{c}{\textbf{StanfordCars}} & \multicolumn{4}{c}{\textbf{Food101}} & \multicolumn{4}{c}{\textbf{DTD}} \\
    \cmidrule(lr){2-5} \cmidrule(lr){6-9} \cmidrule(lr){10-13}
    {\textbf{Shots}} & {2} & {4} & {8} & {16} & {2} & {4} & {8} & {16} & {2} & {4} & {8} & {16}  \\
    \toprule 
    {Tip~\cite{zhang2022tip} } & {57.93} & {61.45} & {62.9} & {66.77} & {77.52} & {77.54} & {77.76} & {77.83} & {49.47} & {53.96} & {58.63} & {60.93} \\
    {CoOp~\cite{zhou2022learning}} & {58.28} & {62.62} & {68.43} & {73.36} & {72.49} & {73.33} & {71.82} & {74.67} & {45.15} & {53.49} & {59.97} & {63.58} \\
    {Tip-F~\cite{zhang2022tip} } & {61.10} & {64.5} & {68.25} & {74.15} & {77.6} & {77.8} & {78.1} & {79.0}  & {53.72} & {57.39} & {62.7} & {65.5} \\
    {Ours} & {\textbf{61.25}} & {\textbf{64.70}} & {\textbf{69.15}} & {\textbf{74.80}} & {\textbf{77.66}} & {\textbf{77.89}} & {\textbf{78.47}} & {\textbf{79.05}} & {\textbf{54.5}} & {\textbf{59.28}} & {\textbf{63.47}} & {\textbf{66.13}} \\ 
\bottomrule
  \end{tabular}}
  \label{tab:cars_food_dtd}
\end{table*}

%

%

\subsection{Ablation studies}
We conduct an ablation study on the novel components of our method in Table.~\ref{tab:ablation}. As expected, adding synthetic images generated by the diffusion model and MMD improves the performance of EuroSAT, SUN397, and UCF101 datasets in low-data settings. MMD seems to be particularly helpful in extremely low data cases (e.g., 2 shot) as evident from Table.~\ref{tab:ablation}. We also provide the ablation of the MMD loss coefficient $\alpha$ in Table.~\ref{tab:ablation_alpha}. It appears that for low-shot cases, higher $\alpha$ works better. 
Ablation on different backbones and the number of generated synthetic images for few-shot classification on ImageNet have been provided in Tab.~\ref{table:backbone} and Tab.~\ref{table:ablation_no_syn_img} respectively.

\begin{table*}[!ht]
  \centering
  \caption{Ablation Study on contributions}
  \vspace{-0.4cm}
  \scalebox{0.7}{
  \begin{tabular}{@{}lccccccccccccc@{}}
    \toprule
    {} & \multicolumn{4}{c}{\textbf{EuroSAT}} & \multicolumn{4}{c}{\textbf{SUN397}} & \multicolumn{4}{c}{\textbf{UCF101}} \\
    \cmidrule(lr){2-5} \cmidrule(lr){6-9} \cmidrule(lr){10-13}
    {\textbf{Shots}} & {2} & {4} & {8} & {16} & {2} & {4} & {8} & {16} & {2} & {4} & {8} & {16}  \\
    \toprule 
    {Tip-F~\cite{zhang2022tip}} & {66.15} & {74.12} & {77.3} & {82.54} & {63.64} & {66.21} & {68.87} & {70.47} & {66.43} &
    {70.55} & {74.01} & {77.03} \\
    {Tip-F + Syn} & {66.8} & {75.93} & {77.3} & {83.64} & {64.4} & {67.4} & {68.9} & {70.88} & {67.9} & {71.7} & {74.20} & {77.12} \\
    {Tip-F + Syn + MMD } & {\textbf{67.03}} & {\textbf{77.37}} & {\textbf{77.5}} & {\textbf{83.64}} & {\textbf{64.6}} & {\textbf{67.47}} & {\textbf{68.93}} & {\textbf{70.9}} & {\textbf{68.57}} & {\textbf{71.76}} & {\textbf{74.12}} & {\textbf{77.24}} \\
\bottomrule
  \end{tabular}}
  \label{tab:ablation}
\end{table*}

\begin{table*}[!ht]
  \centering
  \caption{Ablation on MMD coefficient $\alpha$}
  \vspace{-0.4cm}
  \scalebox{0.8}{
  \begin{tabular}{@{}lccccccccccccc@{}}
    \toprule
    {} & \multicolumn{4}{c}{\textbf{EuroSAT}} & \multicolumn{4}{c}{\textbf{SUN397}} & \multicolumn{4}{c}{\textbf{UCF101}} \\
    \cmidrule(lr){2-5} \cmidrule(lr){6-9} \cmidrule(lr){10-13}
    {\textbf{$\alpha$}} & {2} & {4} & {8} & {16} & {2} & {4} & {8} & {16} & {2} & {4} & {8} & {16}  \\
    \toprule 
    {0 } & {66.086} & {75.93} & {76.45} & {\textbf{83.64}} & {64.46} & {67.45} & {68.91} & {70.88} & {67.9} & {71.76} & {73.56} & {73.77} \\
    {0.01} & {65.86} & {73.5} & {77.08} & {83.02} & {64.302} & {67.45} & {68.9} & {\textbf{70.90}} & {68.27} & {71.76} & {73.51} & {77.21} \\
    {0.1} & {65.29} & {76.64} & {77.38} & {82.75} & {64.31} & {67.45} & {68.63} & {70.61}  & {67.93} & {71.76} & {74.12} & {77.24} \\
    {1} & {\textbf{67.03}} & {\textbf{77.37}} & {\textbf{77.5}} & {83.01} & {\textbf{64.60}} & {67.39} & {\textbf{68.93}} & {70.61} & {\textbf{68.57}} & {71.76} & {\textbf{73.88}} & {\textbf{77.08}} \\
\bottomrule
  \end{tabular}}
  \label{tab:ablation_alpha}
\end{table*}


\begin{table}
  \centering
  \caption{Comparison on Caltech101}
  \vspace{-0.4cm}
  \scalebox{0.8}{
  \begin{tabular}{lccccc}
  \toprule
  {} & \multicolumn{4}{c}{{\bf Caltech101}} \\ 
  \cmidrule(lr){2-5} 
  {\bf Method} & {\bf 2-shot} & {\bf 4-shot} & {\bf 8-shot} & {\bf 16-shot}\\ 
  \midrule
 {Tip~\cite{zhang2022tip} } & {88.44} & {89.39} & {89.83} & {90.18} \\
 {CoOp~\cite{zhou2022learning}} & {87.93} & {89.55} & {90.21} & {91.83} \\
 {Tip-F~\cite{zhang2022tip} } & {89.74} & {90.56} & {91.0} & {91.86} \\
 {Ours} & {\textbf{89.93}} & {\textbf{90.99}} & {\textbf{91.44}} & {\textbf{92.8}} \\  	  
  \bottomrule
  \end{tabular}}
  \label{table:caltech101}
\end{table}

\begin{figure}[t]
  \centering
   \includegraphics[scale=0.55]{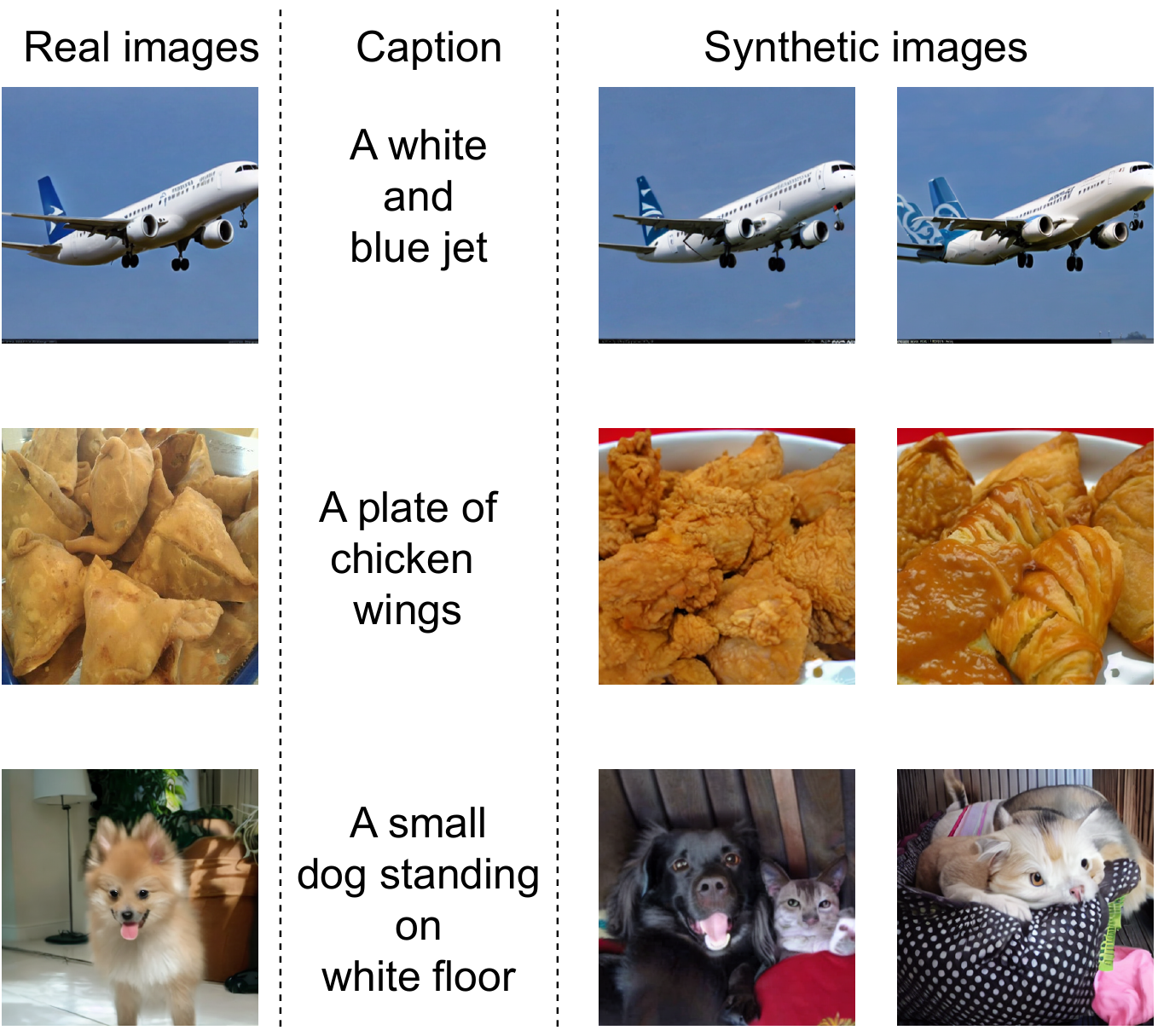}
   \caption{Failure cases: captions for the images are not specific to a particular fine-grained class of images(top row, bottom  row) or are not correctly generated (middle row). Hence, synthetic images are not helpful in classification.}
   \vspace{-0.6cm}
   \label{fig:failure_cases}
\end{figure}

\begin{figure}[t]
  \centering
   \includegraphics[scale=0.65]{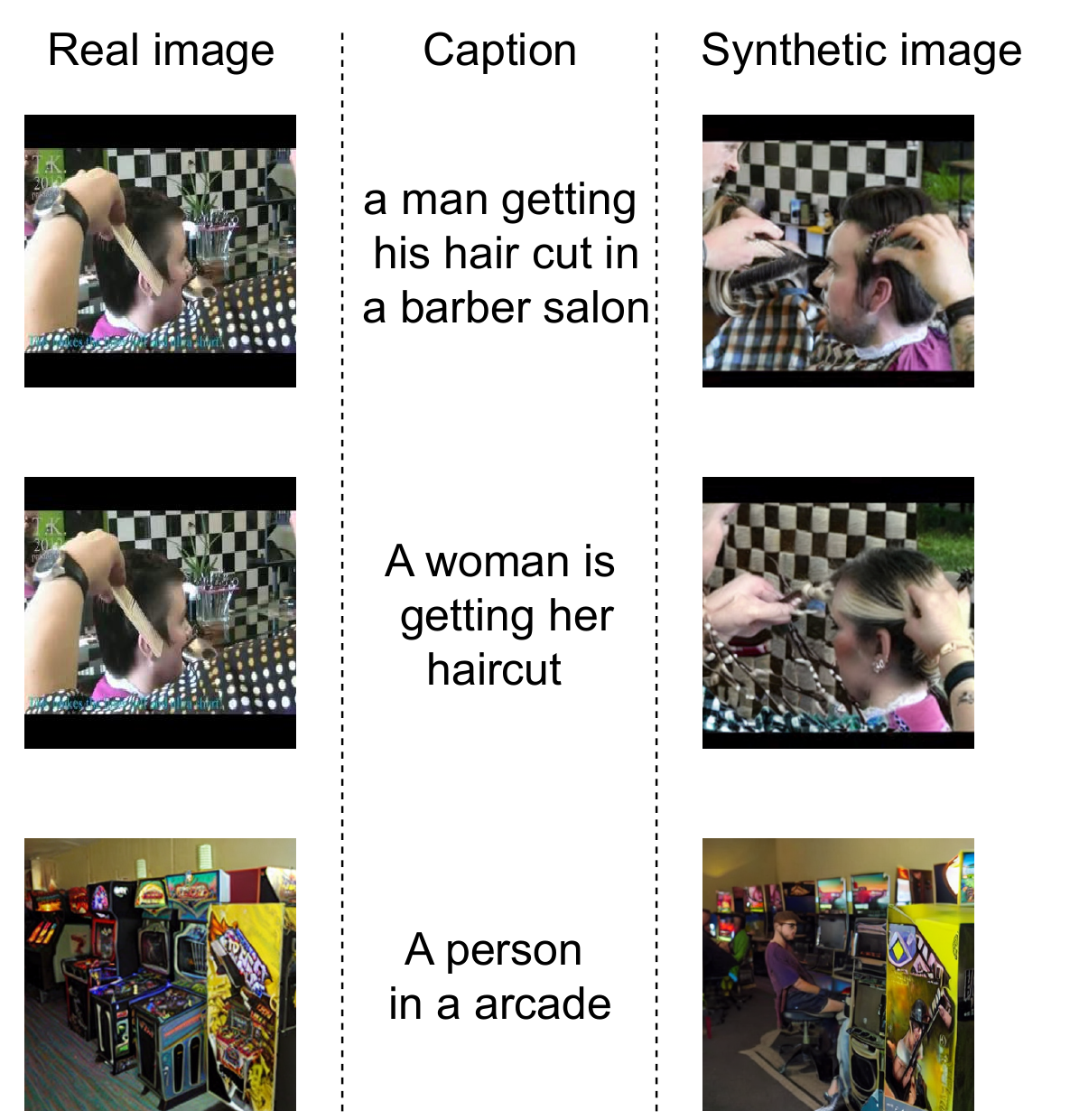}
   \caption{Diverse caption generation. The real image of a man getting a haircut + ``a man getting his hair cut in a barber salon'' when fed to the image-to-image diffusion model produces another image of a man getting a haircut. A real image of a man getting a haircut + ``a woman is getting her haircut'' produces an image of a woman getting a haircut. Therefore, we can do image editing using captions and generate diverse images.}
   \vspace{-0.6cm}
   \label{fig:diverse_caption}
\end{figure}

\begin{figure}[t]
  \centering
   \includegraphics[scale=0.43]{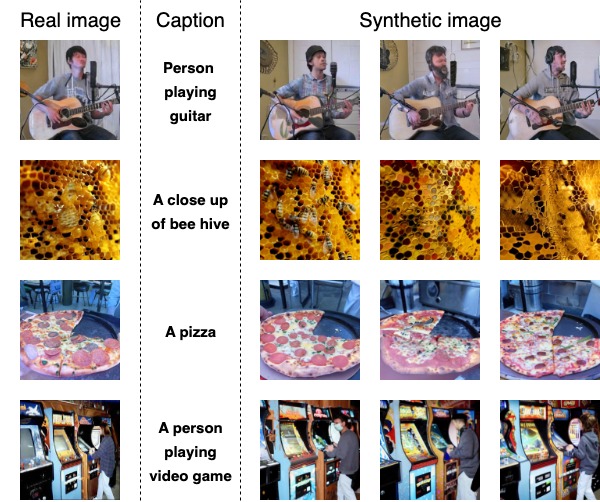}
   \caption{Image to image generation using captions.}
   \label{fig:success_cases}
   \vspace{-0.4cm}
\end{figure}

\subsection{Visualizations}
We show examples where image-to-image generation using caption provides diverse training examples and thus helps provide generalization. E.g., in Fig.~\ref{fig:success_cases} the real image, showing a person playing guitar, and the caption ``person playing guitar'' generates images of different people playing guitar, which helps the model to focus more on ``playing guitar'' (actual class label), than people or background.
Similarly, diverse examples for ``bee-hive'' and ``pizza'' classes are generated by the image and the corresponding captions in Fig.~\ref{fig:success_cases}.


\subsection{Discussions}
Our  method attempts to capture the variations within the class through captions and translate that to generate diverse augmented samples from the training samples using image-to-image diffusion model. For instance, in Fig.~\ref{fig:diverse_caption} (first row) the training image is a picture of a man having his haircut and the corresponding classname is ``haircut'' (from UCF101 dataset). If we provide a caption ``a woman is getting her haircut'' to this image and fed it to the image-to-image diffusion model, it indeed generates an image of a woman having a haircut (second row, right figure). Therefore, such cross-caption based image generation provides diversity in the training set and help generalization. Similarly, in the last row (Fig.~\ref{fig:diverse_caption}) using caption as ``a person in a arcade'' to an image of arcade generates image of an arcade with a person in it, providing more diverse and natural augmented instances.

Since, we are focusing more on cross-captions based image augmentations (Fig.~\ref{fig:teaser_fig}, $I_1C_2$, $I_2C_1$), to ensure more diversity in the augmented samples, we need atleast two images to generate cross-caption based augmentations. Therefore, we are evaluating our approach on 2-shot onwards, ignoring the 1-shot evaluation.

\subsection{Limitations}
Our results indicate that this approach might not be suitable for fine-grained classification, e.g., FGVC-aircraft, Food101 datasets. One potential reason could be that captions are unable to extract the fine-grained details which could be important for fine-grained recognition. E.g., in Fig.~\ref{fig:failure_cases} top row, the airplane is E-195, which has more fine-grained characteristics (e.g., the shape of the plane and wings), than what the caption captures (i.e., ``a white and blue jet''). The synthetic images might confuse with other fine-grained airplane categories and thus degrade performance. Similarly, in the Food101 dataset, the class ``samosa'' (Fig.~\ref{fig:failure_cases} (second row) is miscaptioned as ``plate of chicken wings'', therefore the generated images are not semantically helpful for classifying food items. For the fine-grained pet recognition task, captions are unable to distinguish pet categories, i.e., ``a small dog'' does not differentiate across pet species and therefore our model fails in these cases. We would like to address these limitations in our future work.

\section{Conclusion}
We have proposed Cap2Aug - a data augmentation approach exploiting the image-to-image generative model using captions.
Compared to traditional data augmentation strategies, our proposed augmentation method edits semantic information in the images, captured by image captions.
Our study has shown that the domain gap between real and synthetic images can pose additional challenges. To mitigate this, we have proposed a multi-kernel MMD-based loss function to align synthetic images to real images.
We have validated our approach for long-tail and few-shot classification tasks.
For long-tail classification on the standard ImageNet-LT benchmark, Cap2Aug improves over SOTA methods. Our method outperforms the state-of-the-art approaches on few-shot classification on 11  benchmarks. We have performed ablation studies to justify the contribution of various components of our approach. Finally, we investigate the failure cases and discuss the limitations of our approach.


\section{Acknowledgement}
\label{sec:conclusion}
The authors AR, AS and RC are supported by an ONR MURI grant N00014- 20-1-2787. 


\bibliography{egbib}

\end{document}


\title{Cap2Aug: Caption guided Image to Image data Augmentation \linebreak (Supplementary material)}

\author{First Author\\
Institution1\\
Institution1 address\\
{\tt\small firstauthor@i1.org}
\and
Second Author\\
Institution2\\
First line of institution2 address\\
{\tt\small secondauthor@i2.org}
}
\maketitle

\section*{List of contents}

In the supplementary material, we provide more details of our approach, evaluations, and visualizations. The main sections are as listed below.

\begin{enumerate}
    \item Training details.
    \item Ablation studies.
    \item Visualization of synthetic images.
    \item Failure analysis.
    \item tSNE visualization of the embedding space.
\end{enumerate}

\section{Training details}

In our experiments, we have used an NVIDIA A5000 workstation with 24GB GPU memory. Training details and hyperparameters for our experiments are provided in Table.~\ref{table:hparams}.

\begin{table}[!h]
\begin{center}
\caption{Hyperparamters }
\label{table:hparams}
\scalebox{0.8}{
\begin{tabular}{l|l}
\hline\noalign{\smallskip}
Hyperparameters & Values\\
\noalign{\smallskip}
\hline
\noalign{\smallskip}
\noalign{\smallskip}
Batchsize & 128\\
Learning rate &  0.001 \\
Optimizer &  AdamW \\
Stable diffusion model checkpoint & ``stable diffusion v1-5 model''\\
Residual ratio  & 1 \\
Sharpness ratio & 5.5 \\
strength (diffusion model) & 0.5 \\
guidance scale (diffusion model) & 7.5\\
BLIP-model (caption) & ``blip-image-captioning-large'' \\
\hline 
\end{tabular}}
\end{center}
\end{table}

\section{Ablation studies}

\subsection{ Ablation on our proposed contributions}

In this section, we provide the ablation studies of our proposed methods for OxfordPets, OxfordFlowers, FGVC, StanfordCars, Food101, and DTD datasets in Table.~\ref{tab:ablation_mmd_pets_flowers_fgvc} and Table.~\ref{tab:ablation_mmd_cars_food_dtd} respectively. The results verify that including synthetic images along with MMD loss provides best performance and consistently outperform the state-of-the-art.

\subsection{Ablation on coefficient of MMD loss ($\alpha$)}

In this section, we present the results of ablation studies for the important hyperparameters in our approach. 
Specifically, we observe that when the coefficient of MMD loss ($\alpha$) varies, the classification performance changes.
We show the ablation of $\alpha$ for OxfordPets, OxfordFlowers, FGVC, StanfordCars, Food101, and DTD datasets in Table.~\ref{tab:ablation_alpha_pets_flower_fgvc} and Table.~\ref{tab:ablation_alpha_cars_food_dtd} respectively. 

\section{Visualization of synthetic images}

In this section, we provide some of the visualization of the synthetic images generated by Cap2Aug as shown in Fig.~\ref{fig:supple_examples} for OxfordFlower dataset. The generated images look semantically useful edited versions of the real images and therefore will be considered as useful augmentations. The caption guidance helps to generate diverse synthetic images.

\section{Failure Analysis}

Here we show some of the failure cases incurred by our method as shown in Fig.~\ref{fig:supple_failures}. The images generated for ``falafel'' are not similar to actual ``falafel'' class images, rather it generates images of other food category. The model could be confused while generating the images due to either not having enough information of the ``falafel'' class or it confuses with other food items. 
Similarly, for the cases of ``matted texture'' or ``stop sign'' the failure cases are also shown. 

\section{tSNE visualization of the embedding space}

We have also performed the t-SNE analysis of the features for EuroSAT and OxfordPets dataset in Fig.~\ref{fig:eurosat_tsne} and Fig.~\ref{fig:pets_tsne} respectively. The tSNE plots reveal that the few-shot learner is learning discriminative representations over iterations. 

\begin{table*}[]
  \centering
  \caption{Ablation on MMD coefficient $\alpha$}
  \scalebox{0.97}{
  \begin{tabular}{@{}lccccccccccccc@{}}
    \toprule
    {} & \multicolumn{4}{c}{\textbf{OxfordPets}} & \multicolumn{4}{c}{\textbf{OxfordFlowers}} & \multicolumn{4}{c}{\textbf{FGVC}} \\
    \cmidrule(lr){2-5} \cmidrule(lr){6-9} \cmidrule(lr){10-13}
    {\textbf{$\alpha$}} & {2} & {4} & {8} & {16} & {2} & {4} & {8} & {16} & {2} & {4} & {8} & {16}  \\
    \toprule 
    {0 } & {86.94} & {87.35} & {88.19} & {88.98} & {82.86} & {86.19} & {88.46} & {88.22} & {23.10} & {24.18} & {28.80} & {30.87} \\
    {0.01} & {\textbf{87.22}} & {87.89} & {88.19} & {89.58} & {\textbf{83.06}} & {85.83} & {88.55} & {88.18} & {23.07} & {24.06} & {29.25} & {30.73} \\
    {0.1} & {87.16} & {87.84} & {88.17} & {\textbf{89.58}} & {82.86} & {86.43} & {88.51} & {88.22}  & {23.31} & {\textbf{24.87}} & {29.67} & {33.81} \\
    {1} & {87.08} & {\textbf{87.95}} & {\textbf{88.22}} & {89.56} & {82.01} & {\textbf{86.64}} & {\textbf{88.79}} & {\textbf{89.36}} & {\textbf{23.76}} & {24.63} & {\textbf{29.82}} & {\textbf{34.38}} \\
\bottomrule
  \end{tabular}}
  \label{tab:ablation_alpha_pets_flower_fgvc}
\end{table*}

\begin{table*}[]
  \centering
  \caption{Ablation on MMD coefficient $\alpha$}
  \scalebox{0.97}{
  \begin{tabular}{@{}lccccccccccccc@{}}
    \toprule
    {} & \multicolumn{4}{c}{\textbf{StanfordCars}} & \multicolumn{4}{c}{\textbf{Food101}} & \multicolumn{4}{c}{\textbf{DTD}} \\
    \cmidrule(lr){2-5} \cmidrule(lr){6-9} \cmidrule(lr){10-13}
    {\textbf{$\alpha$}} & {2} & {4} & {8} & {16} & {2} & {4} & {8} & {16} & {2} & {4} & {8} & {16}  \\
    \toprule 
    {0 } & {61.02} & {64.15} & {68.61} & {73.46} & {77.45} & {77.7} & {78.19} & {78.94} & {54.31} & {59.16} & {63.41} & {64.95} \\
    {0.01} & {60.60} & {64.58} & {\textbf{69.01}} & {73.52} & {77.46} & {77.89} & {\textbf{78.45}} & {78.82} & {54.43} & {59.22} & {\textbf{63.47}} & {65.30} \\
    {0.1} & {60.52} & {64.37} & {68.90} & {73.53} & {77.61} & {77.71} & {78.26} & {\textbf{78.97}}  & {\textbf{54.49}} & {\textbf{59.27}} & {63.29} & {\textbf{66.13}} \\
    {1} & {\textbf{61.24}} & {\textbf{64.70}} & {68.63} & {\textbf{74.79}} & {\textbf{77.61}} & {\textbf{77.93}} & {78.30} & {78.75} & {54.25} & {58.62} & {62.88} & {66.07} \\
\bottomrule
  \end{tabular}}
  \label{tab:ablation_alpha_cars_food_dtd}
\end{table*}

\begin{table*}[]
  \centering
  \caption{Ablation Study on contributions}
  \scalebox{0.9}{
  \begin{tabular}{@{}lccccccccccccc@{}}
    \toprule
    {} & \multicolumn{4}{c}{\textbf{OxfordPets}} & \multicolumn{4}{c}{\textbf{OxfordFlowers}} & \multicolumn{4}{c}{\textbf{FGVC}} \\
    \cmidrule(lr){2-5} \cmidrule(lr){6-9} \cmidrule(lr){10-13}
    {\textbf{Shots}} & {2} & {4} & {8} & {16} & {2} & {4} & {8} & {16} & {2} & {4} & {8} & {16}  \\
    \toprule 
    {Tip-F~\cite{zhang2022tip}} & {87.03} & {87.54} & {88.09} & {89.70} & {82.30} & {85.83} & {90.51} & \textbf{{94.80}} & {23.19} & {24.80} & {29.21} & {34.55} \\
    {Tip-F + Syn} 
    & {87.16} & {87.60} & {88.19} & {89.70} 
    & {82.86} & {86.19} & {90.89} & {94.32}
    & {23.23} & {24.80} & {29.50} & \textbf{{34.58}} \\
    {Tip-F + Syn + MMD} & \textbf{{87.33}} & \textbf{{87.92}} & \textbf{{88.20} }& \textbf{{89.73}} & \textbf{{83.06}} & \textbf{{86.64}} & \textbf{{91.44}} & {94.51}  & \textbf{{23.76}} & \textbf{{24.87}} & \textbf{{29.82}} & {34.38} \\  
\bottomrule
  \end{tabular}}
  \label{tab:ablation_mmd_pets_flowers_fgvc}
\end{table*}

\begin{table*}[]
  \centering
  \caption{Ablation Study on contributions}
  \scalebox{0.9}{
  \begin{tabular}{@{}lccccccccccccc@{}}
    \toprule
    {} & \multicolumn{4}{c}{\textbf{StanfordCars}} & \multicolumn{4}{c}{\textbf{Food101}} & \multicolumn{4}{c}{\textbf{DTD}} \\
    \cmidrule(lr){2-5} \cmidrule(lr){6-9} \cmidrule(lr){10-13}
    {\textbf{Shots}} & {2} & {4} & {8} & {16} & {2} & {4} & {8} & {16} & {2} & {4} & {8} & {16}  \\
    \toprule 
    {Tip-F~\cite{zhang2022tip} } & {61.10} & {64.50} & {68.25} & {74.15} & {77.60} & {77.80} & {78.10} & {79.00}  & {53.72} & {57.39} & {62.70} & {65.50} \\
    {Tip-F + Syn} 
    & {61.20} & {64.60} & {68.61} & {74.46}
    & {77.45} & {77.70} & {78.19} & {78.94}
    & {54.31} & {59.16} & {63.41} & {65.95} \\
    {Tip-F + Syn + MMD} & {\textbf{61.25}} & {\textbf{64.70}} & {\textbf{69.15}} & {\textbf{74.80}} & {\textbf{77.66}} & {\textbf{77.89}} & {\textbf{78.47}} & {\textbf{79.05}} & {\textbf{54.50}} & {\textbf{59.28}} & {\textbf{63.47}} & {\textbf{66.13}} \\ 
\bottomrule
  \end{tabular}}
  \label{tab:ablation_mmd_cars_food_dtd}
\end{table*}

\begin{figure*}[]
  \centering
   \includegraphics[scale=0.55]{figure/pets_tsne.pdf}
   \caption{tSNE plot for the image representations across training iterations (left to right) for OxfordPets dataset in 16-shot setting. As training progresses, the representations become more clustered showing the discriminative power of the few-shot classifier.}
   \label{fig:pets_tsne}
\end{figure*}

\begin{figure*}[t]
  \centering
   \includegraphics[scale=0.7]{figure/eurosat_tsne_2.pdf}
   \caption{tSNE plot for the image representations across training iterations (left to right) for EuroSAT dataset in 16-shot setting. As training progresses, the representations become more clustered showing the discriminative power of the few-shot classifier.}
   \label{fig:eurosat_tsne}
\end{figure*}

\begin{figure*}[t]
  \centering
   \includegraphics[scale=0.7]{figure/supple_examples.pdf}
   \caption{Examples of synthetic images generated from real images}
   \label{fig:supple_examples}
\end{figure*}

\begin{figure*}[t]
  \centering
   \includegraphics[scale=0.7]{figure/supple_failure.pdf}
   \caption{Failure cases: The corresponding class names are provided. The synthetic images generated for classes ``Falafel'' from Food101 dataset, ``matted texture'' from DTD dataset and ``Stop Sign'' from Caltech-101 dataset are shown here as some of the failure cases.}
   \label{fig:supple_failures}
\end{figure*}

\clearpage 
{\small
\bibliographystyle{ieee_fullname}
\bibliography{egbib}
}